\documentclass[journal]{IEEEtran}
\usepackage{graphicx,float,amssymb,multirow,pifont,amsmath,scalerel,makecell}
\usepackage{bm,hyperref,xcolor,enumitem,booktabs,arydshln,footnote,url}
\usepackage[caption=false]{subfig}
\usepackage{color}
\usepackage{amsthm,amsmath,amssymb}
\usepackage{mathrsfs}
\usepackage{threeparttable}

\usepackage[subfigure,titles]{tocloft}
\usepackage[comma, numbers]{natbib}
\hypersetup{hypertex=true,
	colorlinks=true,
	linkcolor=blue,
	anchorcolor=blue,
	citecolor=blue}

\begin{document}
\title{SAM-Assisted Remote Sensing Imagery Semantic Segmentation with Object and Boundary Constraints}
\author{Xianping~Ma,
        Qianqian~Wu,
        Xingyu~Zhao,
				Xiaokang~Zhang,
				Man-On~Pun,
				and~Bo Huang
		\thanks{This work was supported in part by the National Natural Science Foundation of China under Grant 42371374 and 41801323, China Postdoctoral Science Foundation under grant 2020M682038 and the Shenzhen Science and Technology Innovation Committee under Grant No. JCYJ20190813170803617. \textit{(Corresponding authors: Man-On Pun and Xiaokang Zhang)}}
		\thanks{Xianping Ma, Qianqian Wu, Xingyu~Zhao and Man-On Pun are with the School of Science and Engineering, The Chinese University of Hong Kong, Shenzhen, Shenzhen 518172, China (e-mails: xianpingma@link.cuhk.edu.cn; qianqianwu@link.cuhk.edu.cn; xzhao911@usc.edu; SimonPun@cuhk.edu.cn).}
		\thanks{Xiaokang Zhang is with the School of Information Science and Engineering, Wuhan University of Science and Technology, Wuhan 430081, China, and also with the Department of Land Surveying and Geo-Informatics, The Hong Kong Polytechnic University, Hong Kong, China.(e-mail: natezhangxk@gmail.com).}
		\thanks{Bo Huang is with the Department of Geography, The University of Hong Kong, Hong Kong, SAR 999077, China (e-mail: hbcuhk@gmail.com).}	
	}
\maketitle
\begin{abstract}
Semantic segmentation of remote sensing imagery plays a pivotal role in extracting precise information for diverse downstream applications. Recent development of the Segment Anything Model (SAM), an advanced general-purpose segmentation model, has revolutionized this field, presenting new avenues for accurate and efficient segmentation. However, SAM is limited to generating segmentation results without class information. Meanwhile, the segmentation map predicted by current methods generally exhibit excessive fragmentation and inaccuracy of boundary. This paper introduces a streamlined framework designed to leverage the raw output of SAM by exploiting two novel concepts called SAM-Generated Object (SGO) and SAM-Generated Boundary (SGB). More specifically, we propose a novel object consistency loss and further introduce a boundary preservation loss in this work. Considering the content characteristics of SGO, we introduce the concept of object consistency to leverage segmented regions lacking semantic information. By imposing constraints on the consistency of predicted values within objects, the object consistency loss aims to enhance semantic segmentation performance. Furthermore, the boundary preservation loss capitalizes on the distinctive features of SGB by directing the model's attention to the boundary information of the object. Experimental results on two well-known datasets, ISPRS Vaihingen and LoveDA Urban, demonstrate the effectiveness and broad applicability of our proposed method. The source code for this work will be accessible at \href{https://github.com/sstary/SSRS}{https://github.com/sstary/SSRS}.
\end{abstract}


\IEEEpeerreviewmaketitle

\section{Introduction}\label{sec:int}
Semantic segmentation of remote sensing imagery entails assigning semantic labels to individual pixels within images captured by various remote sensing sensors. This process is fundamental for diverse downstream geoscience applications, including environmental monitoring \cite{yuan2020deep, cao2022coarse}, land cover mapping \cite{li2022land, xu2023rssformer}, and disaster management \cite{gupta2021deep, khan2021deepsmoke, huang2022evaluation, bo2022basnet}. The primary objective is to accurately partition the image into distinct regions representing different semantic classes, facilitating automated analysis and interpretation of remote sensing data. The advent of deep learning techniques \cite{ronneberger2015u, he2016deep, vaswani2017attention, dosovitskiy2020image} has ushered in numerous high-performance, problem-specific methods in this field. According to the type of network, these methods can be primarily categorized into convolutional neural network (CNN) \cite{diakogiannis2020resunet, FuseNet, hong2020multimodal, hong2023cross}, transformer-based approaches \cite{wang2022novel, xu2021efficient, roy2023multimodal, yang2023gtfn} and hybrid architecture \cite{wang2022unetformer, zhang2022transformer, ma2023unsupervised, wu2023cmtfnet} for remote sensing applications.

Recently, a notable foundation model called Segment Anything Model (SAM) \cite{kirillov2023segment} designed for image segmentation has gained considerable attention from the computer vision community. Trained on an extensive dataset of $11$ million natural images and over one billion masks, SAM distinguishes itself by enabling zero-shot segmentation of new visual objects without prior exposure to them. As the first foundation model for general image segmentation, SAM and its original paper have garnered over $1000$ citations within six months, according to the Google Scholar website. However, two conspicuous limitations hinder the application of SAM to remote sensing image semantic segmentation tasks. Firstly, the generated segmentation masks lack semantic labels. Secondly, due to the disparities between natural images and remote sensing images, SAM's effectiveness in remote sensing tasks is compromised \cite{ren2023segment}. 

Researchers have explored various methods from different perspectives to address these limitations and enhance SAM's performance in remote sensing image semantic segmentation tasks. These approaches include adapting or enhancing SAM \cite{stearns2023segment, ding2023adapting, zhang2023enhancing}, employing few-shot or zero-shot learning strategies \cite{osco2023segment, li2023rs, qi2023self, al2023vision} and leveraging prompt learning techniques \cite{chen2023rsprompter, SAMRS, zhang2023text2seg, sultan2023geosam, julka2023knowledge, yan2023ringmo, wang2023cs}. In particular, SAM-CD \cite{ding2023adapting} directly utilizes FastSAM \cite{zhao2023fast} to extract image features while incorporating the underlying temporal constraints in remote sensing images for change detection supervision. RS-CLIP \cite{li2023rs} employs a curriculum learning strategy to boost the performance of zero-shot classification of remote sensing images using SAM through multiple stages of model finetuning. SAMRS \cite{SAMRS} efficiently generates a large-scale remote sensing segmentation dataset by employing various prompts, thereby significantly expanding existing remote sensing datasets. However, these methods often necessitate the artificial design of complex fine-tuning mechanisms or prompt learning strategies. The former involves modifying the structures of the general semantic segmentation model and training strategies, while the latter requires specific prompts tailored to different datasets. Unfortunately, these factors impede the seamless integration of SAM into the field of remote sensing.

\begin{figure}[t]
\centering
{\includegraphics[width=1.0\linewidth]{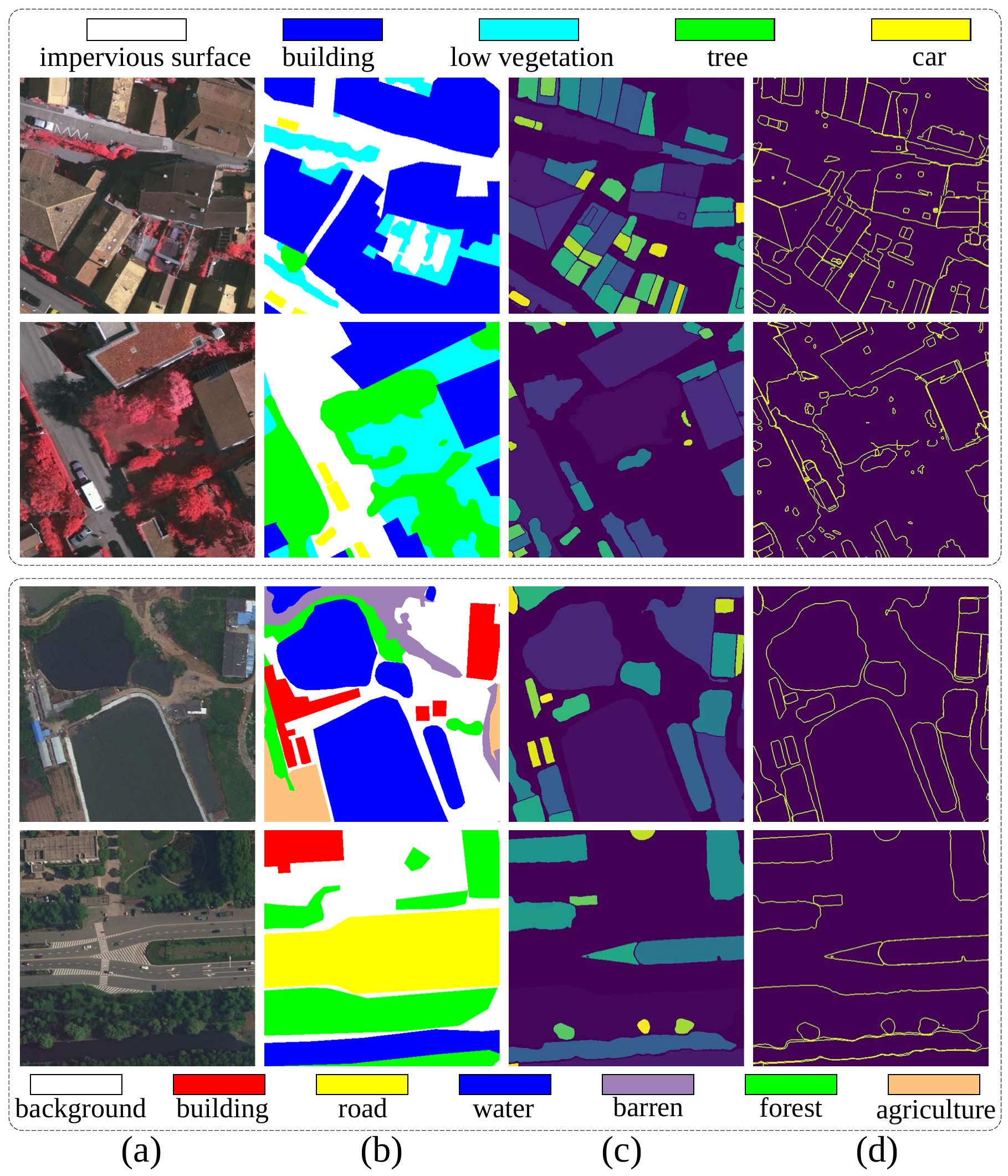}}
\caption{Visual examples. (a) Images, (b) Ground truth, (c) SGO, (d) SGB. The first two rows show samples of size $512 \times 512$ from ISPRS Vaihingen. The last two rows show samples of size $512 \times 512$ from LoveDA Urban. It can be observed that SGO and SGB provide a wealth of detailed information about ground objects from both object and boundary perspectives.}
\label{fig1}
\end{figure}

To cope with the aforementioned challenges, we develop a simple yet effective framework to utilize SAM  from the perspectives of both objects and boundaries. Our observations reveal that SAM encounters challenges in accurately segmenting remote sensing images due to disparities between remote sensing images and natural images. Nevertheless, SAM exhibits proficiency in recognizing \emph {objects}, as evident in Fig.~\ref{fig1}. It shows that SAM-Generated Object (SGO) and SAM-Generated Boundary (SGB) can provide detailed object and boundary information. To fully leverage their potential while minimizing modifications to the general semantic segmentation model, we propose a novel loss function, namely object consistency loss and further introduce a boundary preservation loss to aid in model training. 

For the object consistency loss, it is observed that the regions in SGO, as shown in Fig.~\ref{fig1} (c), are actually objects without semantic information. This insight motivates the design of the object consistency loss, aiming to ensure consistency within a segmented object. Furthermore, we introduce the boundary preservation loss \cite{bokhovkin2019boundary} to encourage the semantic segmentation model to consider better segment boundaries based on the detailed SGB information.  As both loss functions take the semantic segmentation output from the general semantic segmentation model as inputs, there is no need for additional segmentation heads at the tail of the decoder. By harnessing a simple and direct utilization of SGO and SGB, the foundational visual knowledge embedded in SAM contributes to enhancing the semantic segmentation performance of remote sensing images. This approach holds the potential for direct implementation across various tasks and semantic segmentation models involving the combination of SAM and remote sensing. The contributions of this work can be summarized as follows:
\begin{itemize}
\item A streamlined framework is proposed to efficiently leverage two novel concepts called SGO and SGB for remote sensing image semantic segmentation, which emphasizes the value and effectiveness of the \emph {raw} output of SAM. Notably, our approach stands out by not necessitating specific designs for the semantic segmentation model, training strategy, or pseudo-label generation, in contrast to other existing SAM-based methods employed in remote sensing;

\item We propose a novel object consistency loss by restraining the consistency of pixels within an object, and further introduce the boundary preservation loss to assist with model optimization, making full use of the raw output of SAM. This is achieved without relying on semantic information, focusing on two crucial perspectives: object and boundary. To our best knowledge, this work is the first to introduce object and boundary constraints into semantic segmentation tasks to refine the segmentation results by directly utilizing the raw output of SAM without the need for additional class prompts;

\item Extensive experiments on two well-known, publicly available remote sensing datasets, ISPRS Vaihingen and LoveDA Urban, and four representative semantic segmentation models confirm that the proposed approach can be widely adapted to semantic segmentation tasks of different datasets and different general models. We believe that this approach has the potential to significantly expand the application of SGO and SGB, unlocking the full capabilities of large models like SAM in remote sensing image semantic segmentation tasks.
\end{itemize}

The remainder of this paper is organized as follows. Section~\ref{sec:rel} first reviews the related works of SAM in different fields. After that, Section~\ref{sec:met} presents the proposed method in detail, whereas Section~\ref{sec:exp} provides experimental results and discussions. Finally, the conclusion is given in Section~\ref{sec:con}.

\begin{figure*}[t]
\centering
{\includegraphics[width=0.95\linewidth]{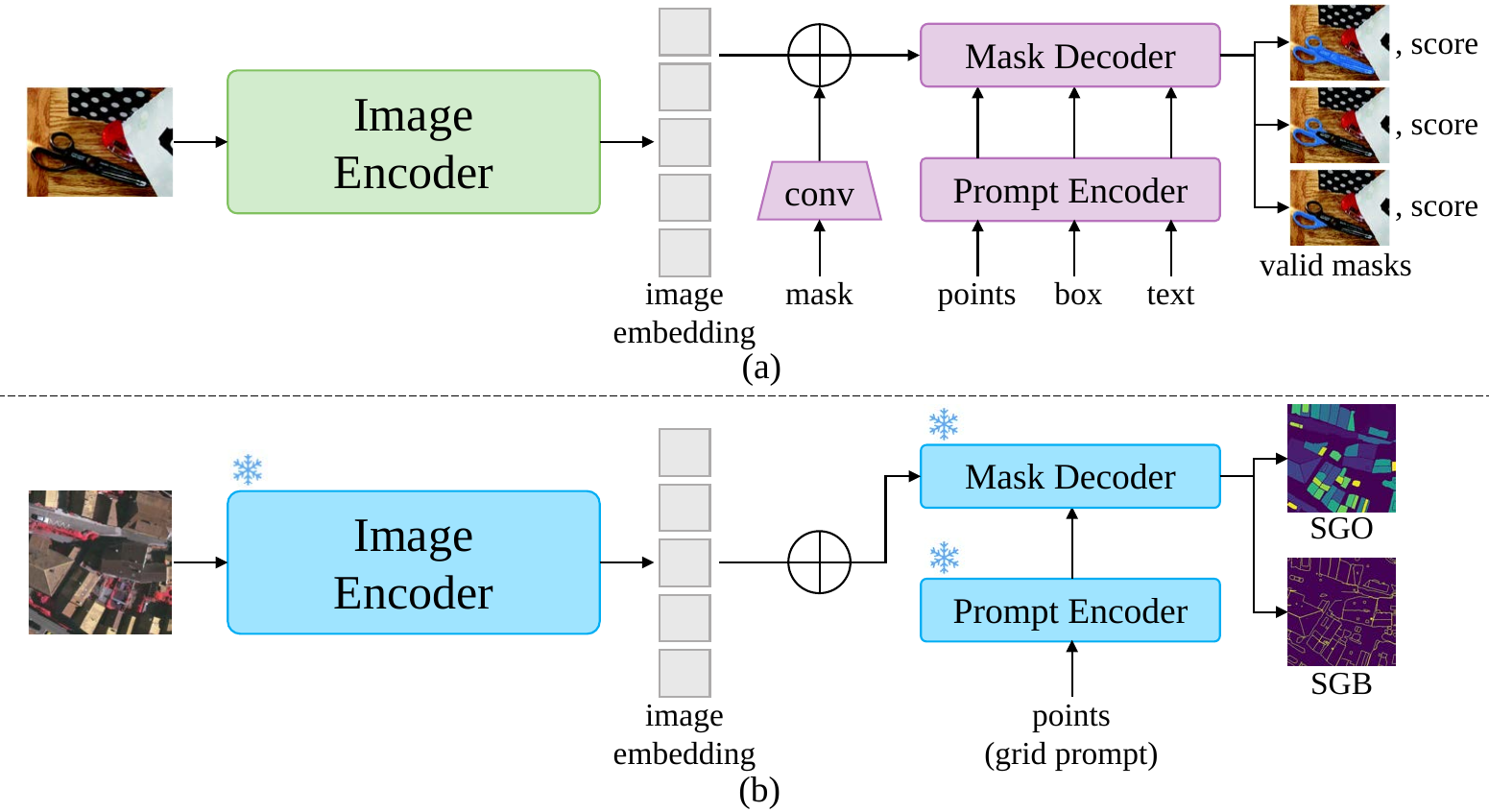}}
\caption{Comparison of frameworks between SAM and our approach. (a) Schematic of the SAM framework \cite{kirillov2023segment}. (b) Schematic of the proposed SAM-based pre-processing approach. All parameters of SAM are frozen, and SGO and SGB are generated with only points prompt.}
\label{fig1.1}
\end{figure*}

\section{Related Works}\label{sec:rel}
\subsection{Segment Anything Model}
The Segment Anything Model (SAM) \cite{markus2014use}, introduced by Meta AI, is a large vision transformer \cite{dosovitskiy2020image}-based model trained on a substantial visual corpus. It is a foundation model aiming to address specific downstream image segmentation tasks. One of the challenges in applying deep neural networks to real-world semantic segmentation applications is the need for large amounts of well-annotated training data. SAM effectively addresses this issue by enabling zero-shot generalization to {\em unseen} images and objects based on user-provided prompts. The framework of SAM has three main components: an image encoder, a prompt encoder, and a mask decoder as shown in Fig.~\ref{fig1.1} (a). The image encoder utilizes a vision transformer-based approach for extracting image features. The prompt encoder incorporates user interactions for segmentation tasks. Different types of prompts are supported, including mask, points, box, and text prompts. The mask decoder comprises transformer layers with dynamic mask prediction heads and an Intersection-over-Union (IoU) score regression head. It maps the encoder embedding, prompt embeddings, and an output token to a mask. For user convenience, SAM offers a set of Application Programming Interfaces (APIs) with which segmentation masks can be obtained in just a few lines of code. Different segmentation mode options, e.g., fully automatic, bounding box, and point mode, are supported for different prompts in APIs.

Currently, SAM has made significant strides in diverse fields. In the realm of medical image processing, \citep{zhang2023input} introduced a straightforward image enhancement method by combining SAM-generated masks with raw images. Additionally, nnSAM \cite{li2023nnsam} integrated the encoder of UNet with the pre-trained SAM encoder, harnessing the feature extraction capabilities of this large foundational vision model. Furthermore, \citep{jiang2023segment, huang2023push} explored SAM's capacity to generate pseudo labels. Through these distinct technical approaches, these methods have propelled the application and development of SAM across various domains.

\subsection{Segment Anything Model in Remote Sensing}
The fundamental distinction between natural images and remote sensing images lies in their acquisition and context, encompassing factors such as acquisition methods, spectral and spatial resolutions, object scale and coverage, and content complexity \cite{li2019deep, zheng2020foreground, ma2023unsupervised}. SAMRS \cite{SAMRS} introduced a prompt known as Rotated Bounding Box (R-Box) to guide SAM segmentation, subsequently generating a comprehensive remote sensing segmentation dataset. This groundbreaking work has paved the way for the integration of large-scale models and the utilization of big data in the field of remote sensing. On a parallel track, Text2Seg \cite{zhang2023text2seg} devised a framework employing multiple foundational models to guide semantic segmentation of remote sensing images through text prompts. However, prompt learning techniques explored in \cite{chen2023rsprompter, SAMRS, zhang2023text2seg, sultan2023geosam, julka2023knowledge} necessitate careful selection based on the specific dataset characteristics, limiting the general applicability of SAM. Meanwhile, few-shot or zero-shot methods \cite{li2023rs, qi2023self, al2023vision} demonstrate promising adaptability for remote sensing tasks, though their sensitivity to additional fine-tuning techniques remains a notable consideration \cite{osco2023segment}. More importantly, the absence of semantic information in SGO confines existing methods to binary classification tasks \cite{li2023rs, ding2023adapting, sultan2023geosam}, or to multiple classification predictions via class-specific prompts \cite{stearns2023segment, ren2023segment, li2023rs}, which undoubtedly limits the progression of SAM in remote sensing. In light of the aforementioned discussions, it is imperative to design an accessible and user-friendly framework for leveraging SGO and SGB in semantic segmentation tasks with multiple classes in remote sensing.

\begin{figure*}[t]
\centering
{\includegraphics[width=0.90\linewidth]{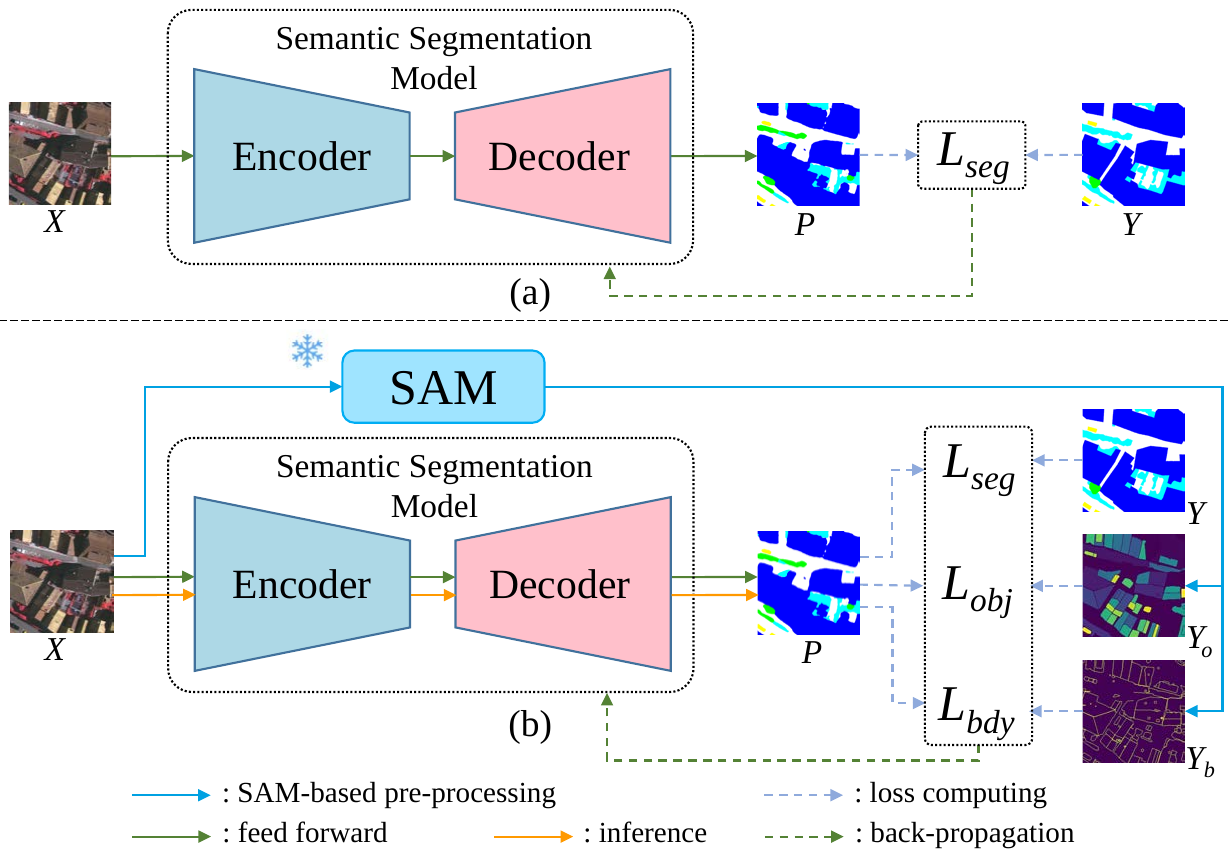}}
\caption{(a) The schematic diagram of the general semantic segmentation model. (b) The schematic diagram of the proposed method. In our method, the image $X$ is fed into the semantic segmentation model that generates the segmentation output $P$ before the segmentation loss $L_{seg}$ is computed by segmentation output and ground truth $Y$, which is used for back-propagation to update the model. Meanwhile, two additional loss functions $L_{obj}$ and $L_{bdy}$ are computed by SGO $Y_{o}$ and SGB $Y_{b}$ that will also be used for model optimization together with $L_{seg}$.}
\label{fig2}
\end{figure*}

\subsection{Object-based methods in Remote Sensing}
The semantic segmentation of remote sensing imagery generally revolves two primary approaches: pixel-based and object-based methods \cite{duro2012comparison} that leverage distinct scales for feature learning and final category prediction. There have been many works explored in integrating object-based concepts into various remote sensing tasks \cite{zhang2018object, martins2020exploring, zhang2018object, zhang2021style, li2022temporal, rittenhouse2022object}. In particular, OCNN \cite{zhang2018object} proposed the first object-based CNN framework to perform land use classification in the complicated scenarios. SDNF \cite{mi2020superpixel} and ESCNet \cite{zhang2021escnet} constructed semantic segmentation networks based on superpixels. The former introduced a superpixel-enhanced region module to mitigate noise and strengthen ground object edges, while the latter proposed an adaptive superpixel merging module, optimizing the model towards objects through handling high-dimensional features. OBIC-GCN \cite{zhang2021object} delved into relationships between objects leveraging graph convolutional networks \cite{henaff2015deep}. However, a notable issue with these methods is that they predominantly focus on extracting object-based features. This approach necessitates the creation of specialized learnable modules to assist in achieving accurate semantic segmentation, thereby elevating the complexity and potential instability of the implementation process \cite{pan2021simplified}. In contrast, the proposed approach integrates seamlessly with existing semantic segmentation models, allowing for straightforward optimization. This method efficiently leverages the raw output of SAM, which contains detailed object information, enhancing overall model performance.

\section{Methodology}\label{sec:met}
The schematic diagram of the proposed framework is depicted in Fig.~\ref{fig2}. 
Fig.~\ref{fig2} (a) showcases the conventional approach to semantic segmentation, where the input image is fed into the semantic segmentation model to generate segmentation output. This output is then utilized to calculate the segmentation loss, followed by model updates through back-propagation. In contrast, our method, as shown in Fig.~\ref{fig2} (b), incorporates an extra stage that employs the SAM. Specifically, we directly create SGO and SGB using the SAM as provided by Meta AI \cite{kirillov2023segment}. These outputs play a crucial role in computing the object consistency loss and boundary preservation loss, respectively, thereby contributing to the model's training. In this section, we will provide detailed explanations of SAM-based pre-processing, network training, and the associated loss functions.

\begin{figure}[t]
\centering
{\includegraphics[width=0.95\linewidth]{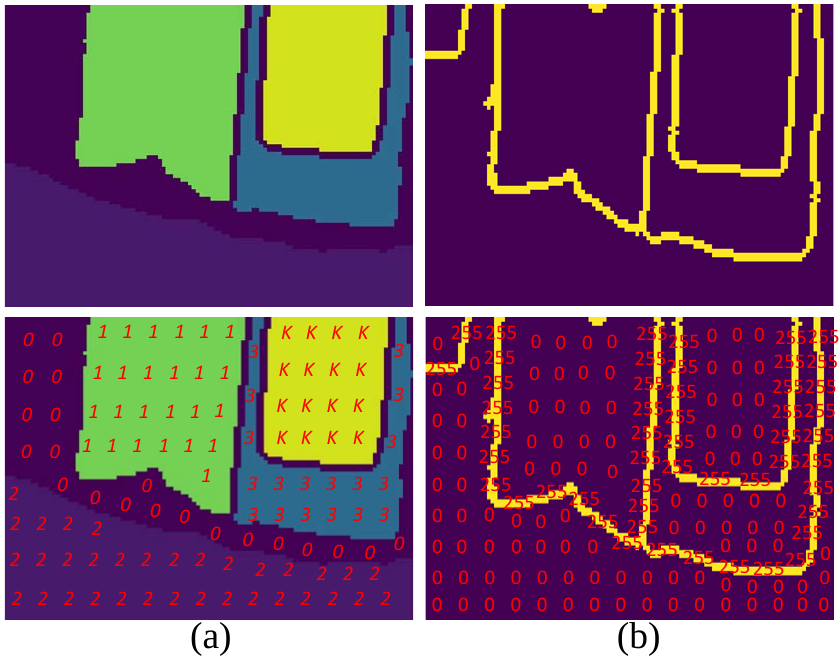}}
\caption{(a) SGO and its diagram with pixel values, (b) SGB and its diagram with pixel values. }
\label{fig2_1}
\end{figure}

\subsection{SAM-based Pre-processing}
The schematic of the proposed SAM-based pre-processing approach is presented as Fig.~\ref{fig1.1} (b). SAM provides a grid prompt technique to automatically process images. Given the input remote sensing image denoted by $X\in \mathbb{R}^{H\times W \times 3}$, SAM can generate segmentation masks across the entire image at all plausible locations in the grid prompt setting \cite{zhang2023input}. In this study, we refer to the segmentation mask as the \emph {object}, considering each segmentation mask as an individual enclosed region that can be regarded as an object. The generated objects are then stored in a list where we set a threshold of $K$ to limit the maximum number of objects in $X$. Meanwhile, we also establish a threshold $S$ to limit the number of pixels that a single object can contain, effectively filtering out very small segmentation masks. Consequently, an SGO denoted by $Y_{o}\in \mathbb{R}^{H\times W}$ can be obtained, where the value of each pixel falls within the range of $[0, K]$. Pixels not segmented as objects, as well as boundaries, are assigned a value of zero, while the objects in $Y_{o}$ are indexed by an identifier denoted as $i$, where $i \in [1, K]$. The data organization of SGO is presented in Fig.~\ref{fig2_1} (a). Concurrently, a boundary prior map is derived from SGO. This process involves outlining the exterior boundaries of each object within the list and merging these boundaries to produce a comprehensive boundary prior map, namely SGB, denoted by $Y_{b}\in \mathbb{R}^{H\times W}$. Unless specified otherwise, the identifier for the boundary pixels in $Y_{b}$ is set to $255$, while others are set to $0$ in the sequel as shown in Fig.~\ref{fig2_1} (b). Visual examples of SGO and SGB are illustrated in Fig.~\ref{fig1} (c) and (d).

\subsection{Network Training}
The classical encoder-decoder networks, e.g., UNetformer \cite{wang2022unetformer}, are widely used in semantic segmentation methods. In this work, we use them as the semantic segmentation model in the proposed framework. Given the input image $X$, the semantic segmentation model produces the predicted segmentation output denoted by $P\in \mathbb{R}^{H\times W \times C}$ where $C$ is the number of categories of ground objects. The learning objective is to minimize the following cross-entropy-based segmentation loss with respect to the parameters of the semantic segmentation model:
\begin{eqnarray}
L_{seg}=-\sum_{\mathcal H, \mathcal W} \sum_{c \in C}Y^{(\mathcal H, \mathcal W,c)}\mathrm{log} (P^{ (\mathcal H, \mathcal W,c)}),\label{eq1}
\end{eqnarray}
where $Y$ stands for the ground truth.

Given that SGO and SGB are solely utilized for computing loss functions, the proposed framework requires no additional modifications or adjustments to the network and training strategies. Thus, the learning objective for our method is to minimize the following composite loss function:
\begin{eqnarray}
L_{total}=L_{seg} + \lambda_{o} L_{obj} + \lambda_{b} L_{bdy},\label{eq2}
\end{eqnarray}
where $L_{obj}$ and $L_{bdy}$ stand for the object consistency loss and boundary preservation loss, respectively. Furthermore, $\lambda_{o}$ and $\lambda_{b}$ are two weighting coefficients to balance the three losses.

\begin{figure}[t]
\centering
{\includegraphics[width=0.95\linewidth]{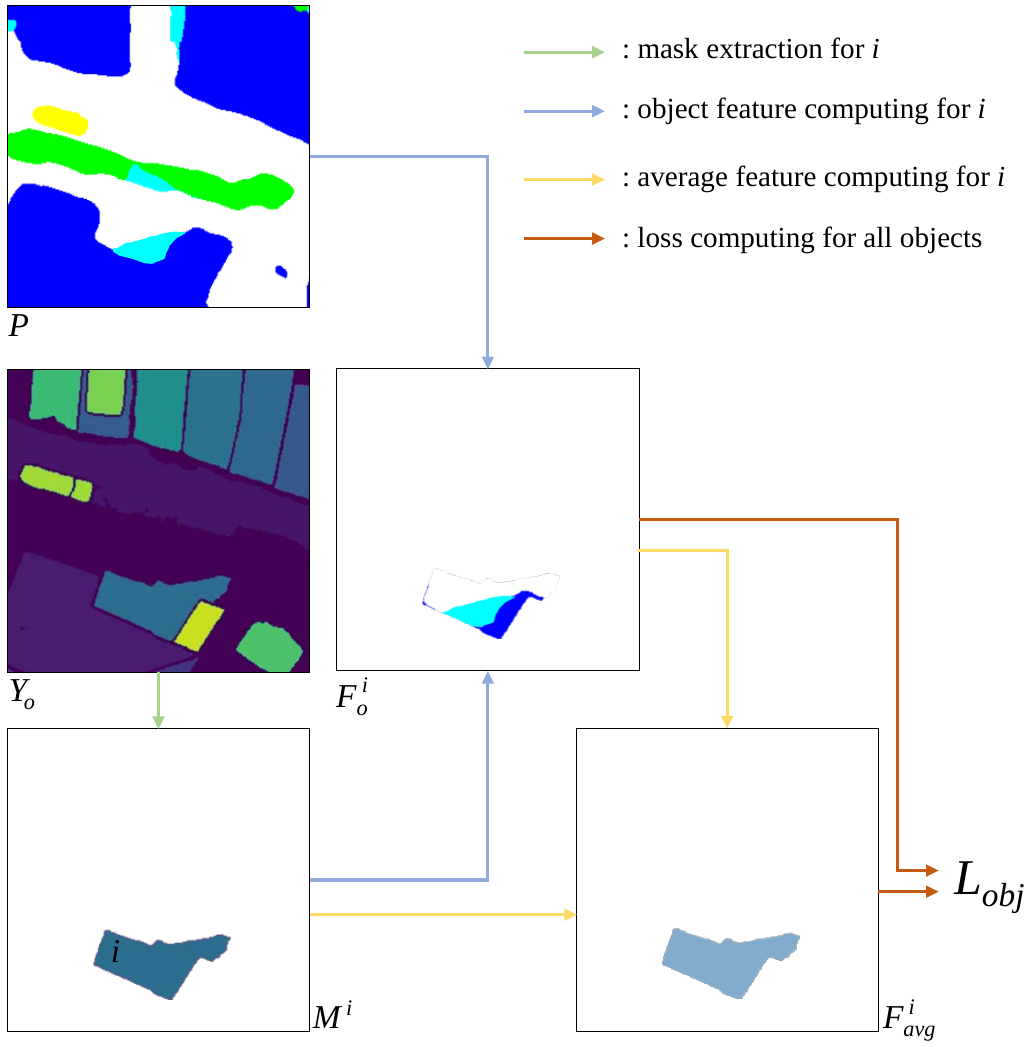}}
\caption{The flow chart for computing object consistency loss.}
\label{fig2_2}
\end{figure}

\subsubsection{Object Consistency Loss}
The object consistency loss is designed to maintain pixel consistency within the objects in a given input image. Given the input $X$, the output of the semantic segmentation model is denoted as $P$. To compute the object consistency loss, we iterate through all the objects in $Y_{o}$. The data flow is presented in Fig.~\ref{fig2_2}. For each object, we first extract its mask $M^{i}$, i.e. the region in $Y_{o}$ where the pixel value equals $i$. The object feature is then obtained by:
\begin{eqnarray}
F^{i}_{o}=P \odot M^{i}, \label{eq3}
\end{eqnarray}
where $\odot$ stands for the Hadamard product. The object feature $F^{i}_{o}$ denotes the model prediction filtered based on the area of the $i$-th object. Next, we can calculate the average feature of the object as:
\begin{eqnarray}
F^{i}_{avg}=\frac{\mathcal{G} (F^{i}_{o})}{N^{i} + 1} \odot M^{i}, \label{eq4}
\end{eqnarray}
where $\mathcal{G}$ calculates the sum of all pixels in the spatial dimension and reshapes to its original shape while $N^{i}$ is the number of points in the $i$-th object. An extra one is added to avoid the denominator being zero. $F^{i}_{avg}$ represents the expected mean value of all the pixels in the $i$-th object. Thus, we can compute the object consistency loss $L_{obj}$ for all objects as:
\begin{eqnarray}
L_{obj}=\sum_{i=1}^{K} \mathcal{MSE} (F^{i}_{o}, F^{i}_{avg}), \label{eq5}
\end{eqnarray}
where $\mathcal{MSE}(\cdot)$ is the mean squared error function. 

Clearly, the proposed $L_{obj}$ directly leverages the regions generated by SAM, fully utilizing the detailed segmentation mask information in SGO.

\subsubsection{Boundary Preservation Loss}
Previous studies \cite{marmanis2018classification, liu2018ern, bokhovkin2019boundary} have demonstrated that incorporating edge constraints can effectively enhance the performance of semantic segmentation models in remote sensing tasks. Our observations indicate that SGO inherently contains highly detailed boundary information, as depicted in Fig.~\ref{fig1} (d). To leverage this boundary information, we set the boundary to $0$ in $Y_{o}$ and generate SGB, denoted as $Y_{b}$. In this work, the boundary metric ($BF_{1}$) \cite{bokhovkin2019boundary}, capable of directly computing boundary preservation loss from the segmentation output $P$ of the semantic model, is employed to evaluate the precision of boundary detection. The boundary preservation loss $L_{bdy}$ is given by:
\begin{eqnarray}
L_{bdy}=1 - BF_{1}, \label{eq6}
\end{eqnarray}
where $BF_{1}$ is defined as:
\begin{eqnarray}
BF_{1}=2 \times \frac{p_{b}r_{b}}{p_{b} + r_{b}}, \label{eq7}
\end{eqnarray}
with $p_{b}$ and $r_{b}$ being the precision and recall of the boundary that can comprehensively evaluate the accuracy of the boundary detection results from $P$ and $Y_{b}$ \cite{bokhovkin2019boundary}.

Finally, the overall objective function employed in training the semantic segmentation model is given by Eq.~\eqref{eq2}, which sums the semantic segmentation loss $L_{seg}$, the object consistency loss $L_{obj}$, and the boundary preservation loss $L_{bdy}$.

\begin{table*}\footnotesize
	\centering
	\caption{Experimental results on the ISPRS Vaihingen dataset. We present the OA of five foreground classes and three overall performance metrics.  The accuracy of each category is presented in the F1/IoU form. Bold values are the best.}
	  \renewcommand\arraystretch{1.2}
		\begin{tabular}{ccccccccc}
			\hline
			\textbf{Method} & \textbf{Backbone} &  \textbf{impervious surface} & \textbf{building} & \textbf{low vegetation}  & \textbf{tree}  & \textbf{car}  & \textbf{mF1}    & \textbf{mIoU} \\
			\hline
			ABCNet  \cite{li2021abcnet}  &ResNet-18& 89.78/81.45 & 94.30/89.21 & 78.49/64.59 & \textbf{90.08/81.95} & 74.05/58.80 & 85.34 & 75.20 \\
			ABCNet+SAM   &ResNet-18& \textbf{92.02/85.21} & \textbf{95.97/92.25} & \textbf{79.90/66.53} & 90.06/81.92 & \textbf{78.52/64.64} & \textbf{87.29} & \textbf{78.11} \\
			\hline
			CMTFNet      \cite{wu2023cmtfnet}   &ResNet-50& 92.53/86.09 & 96.95/94.09 & 79.98/66.64 & 90.22/82.19 & 89.87/81.60 & 89.91 & 82.12 \\
			CMTFNet+SAM      &ResNet-50& \textbf{92.98/86.88} & \textbf{97.05/94.26} & \textbf{80.58/67.48} & \textbf{91.19/83.81} & \textbf{90.98/83.44} & \textbf{90.56} & \textbf{83.18}\\
      \hline
			UNetformer \cite{wang2022unetformer}    &ResNet-18& 92.33/85.76 & 96.25/92.78 & 80.47/67.33 & 90.85/83.22 & 89.35/80.75 & 89.85 & 81.97 \\
			UNetformer+SAM &ResNet-18& \textbf{92.79/86.54} & \textbf{96.74/93.69} & \textbf{80.85/67.86} & \textbf{91.17/83.77} & \textbf{90.43/82.53} & \textbf{90.40} & \textbf{82.88} \\
			\hline
			FTUNetformer \cite{wang2022unetformer}   &SwinTrans-base& 93.41/87.64 & 96.92/94.02 & 81.53/68.82 & 90.91/83.33 & 88.46/79.31 & 90.24 & 82.62 \\
			FTUNetformer+SAM &SwinTrans-base&\textbf{93.73/88.20} & \textbf{97.19/94.53} & \textbf{81.77/69.16} & \textbf{91.62/84.53} & \textbf{91.09/83.64} & \textbf{91.08} & \textbf{84.01} \\
			\hline
	\end{tabular}\label{tab:vlist}
\end{table*}

\section{Experiments and Discussion}\label{sec:exp}
\subsection{Datasets}
\subsubsection{ISPRS Vaihingen} The ISPRS Vaihingen dataset comprises $16$ true orthophotos with very high-resolution, averaging $2500\times2000$ pixels. Each orthophoto encompasses three channels, namely Near-Infrared, Red, and Green (NIRRG) of $9$ cm ground sampling distance. This dataset includes five foreground classes, namely {\em impervious surface}, {\em building}, {\em low vegetation}, {\em tree}, {\em car}, and one background class ({\em clutter}). The $16$ orthophotos are divided into a training set of $12$ patches and a test set of $4$ patches. The training set comprises orthophotos of index numbers $1, 3, 23, 26, 7, 11, 13, 28, 17, 32, 34, 37$, while the test set $5, 21, 15, 30$.

\subsubsection{LoveDA Urban} The LoveDA dataset contains two scenes, namely Urban and Rural. Considering the diverse distribution of ground objects, we selected the LoveDA Urban scene for our experiments. The LoveDA Urban comprises $1833$ high-resolution optical remote sensing images, each with the size of $1024\times 1024$ pixels. The images provide three channels, namely Red, Green, and Blue (RGB), with a ground sampling distance of $30$ cm. The dataset encompasses seven landcover categories, including {\em background}, {\em building}, {\em road}, {\em water}, {\em barren}, {\em forest}, and {\em agriculture} \cite{wang2021loveda}. These images were collected from three cities in China (Nanjing, Changzhou, and Wuhan). The $1833$ images are divided into two parts, with $1156$ images for training and $677$ images for testing. Specifically, the training set contains images indexed from $1366$ to $2521$, while the test set spans images from $3514$ to $4190$.

The two datasets differ in sampling resolution, ground object categories, and label accuracy. Notably, the ISPRS Vaihingen exhibits higher sampling accuracy, a reduced number of categories, and more precise ground truth. In contrast, the LoveDA Urban presents the opposite characteristics. Conducting experiments on both datasets provides substantial evidence regarding the effectiveness of our framework. Throughout the training and testing processes, a sliding window is employed for dynamically assembling training batches, enabling the processing of large images without pre-splitting. The sliding window size is set to $256\times 256$, with a stride of $256$ during training and an adjusted stride of $32$ during testing. The use of a smaller stride in testing effectively mitigates border effects by averaging the prediction outcomes within overlapping regions \cite{audebert2018beyond, ma2022crossmodal}.

\begin{figure*}[hbp]
\centering
{\includegraphics[width=0.85\linewidth]{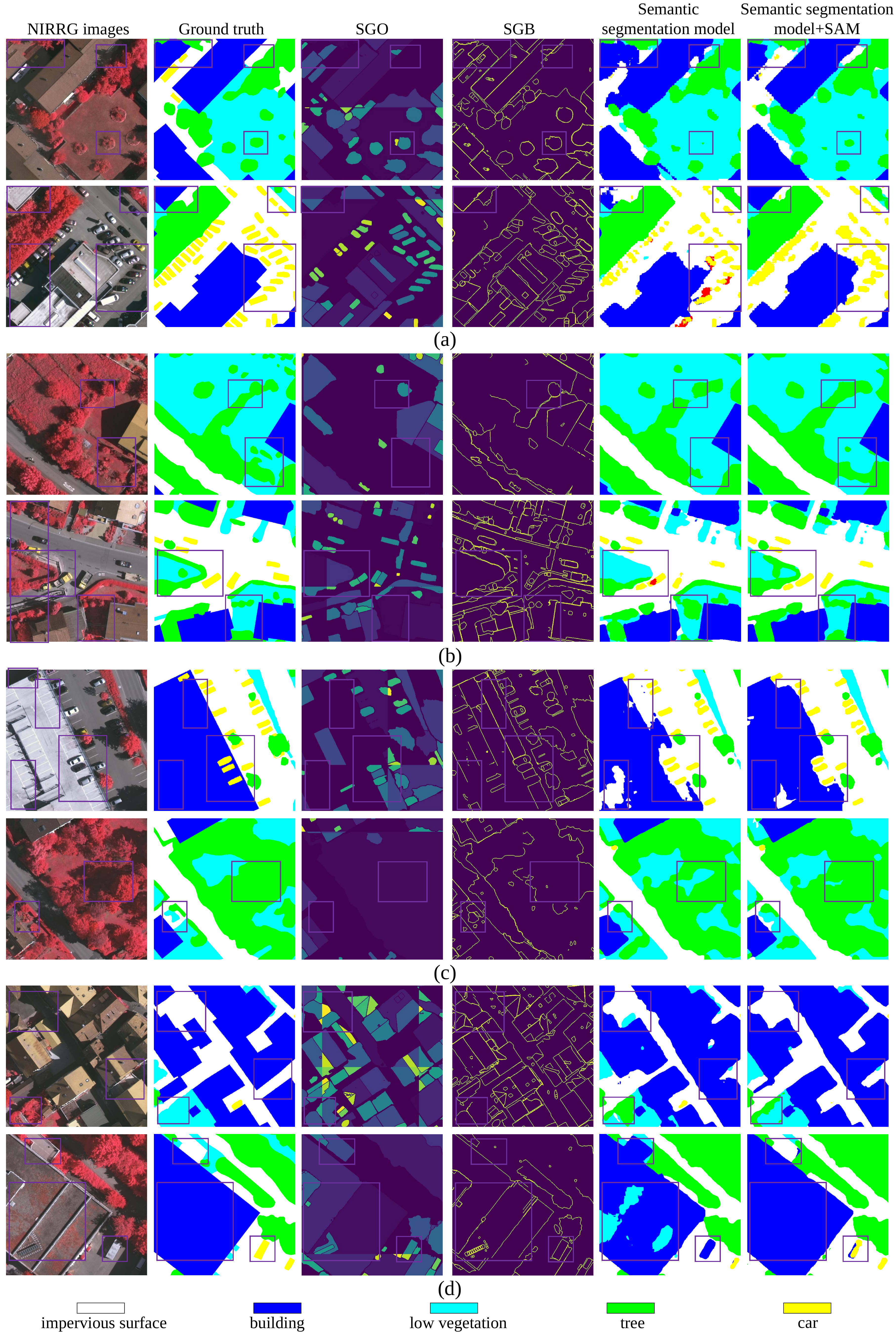}}
\caption{Qualitative performance comparisons on the ISPRS Vaihaigen with the size of $512 \times 512$. (a) ABCNet, (b) CMTFNet, (c) UNetformer, (d) FTUNetformer. We showcase two samples for each model. Some purple boxes are added to highlight the differences.}
\label{fig3}
\end{figure*}

\subsection{Evaluation Metrics}
To assess the segmentation performance of the proposed framework, the mean F1 score (mF1) and the mean Intersection over Union (mIoU) are employed in our experiments. These widely used statistical indices allow for fair comparisons between the performance of our method and state-of-the-art baseline methods. In particular, we compute mF1 and mIoU of the five foreground classes for the ISPRS Vaihingen. The class labeled as {\em Clutter} or {\em Background} is treated as a cluttered and sparse class, and thus, performance statistics are not calculated for either class \cite{diakogiannis2020resunet, wu2023cmtfnet}. In the case of LoveDA Urban, all seven categories are considered in our experiments. F1 and IoU metrics are computed for each class identified by the index $c$ using the following formulas:
\begin{eqnarray}
\text{F1} = 2\times\frac{p_{c}r_{c}}{p_{c}+r_{c}},\label{eq8}
\end{eqnarray}
\begin{eqnarray}
\text{IoU} = \frac{TP_{c}}{TP_{c}+FP_{c}+FN_{c}},\label{eq9}
\end{eqnarray}
where $TP_{c}$, $FP_{c}$, and $FN_{c}$ are true positives, false positives, and false negatives for the $c$-th class, respectively. Furthermore, $p_{c}$ and $r_{c}$ are given by:
\begin{eqnarray}
p_{c}&=&\frac{TP_{c}}{TP_{c}+FP_{c}},\\\label{eq19}
r_{c}&=&\frac{TP_{c}}{TP_{c}+FN_{c}}.\label{eq20}
\end{eqnarray}
Upon computing F1 and IoU for the main classes according to the definitions above, we derive their mean values, denoted as mF1 and mIoU, respectively.

\subsection{Implementation details}
The experiments were conducted using PyTorch on a single NVIDIA GeForce RTX 4090 GPU equipped with 24GB RAM. To generate SGO and SGB, we utilize the interface offered by Meta AI, which involves three pertinent parameters, namely ``crop\_nms\_thresh", ``box\_nms\_thresh", and ``pred\_iou\_thresh" whose definitions can be found in Meta AI documentation\footnote{Meta AI: \url{https://github.com/facebookresearch/segment-anything/tree/main/segment_anything}}. In our experiments, these values were configured to $0.5$, $0.5$, and $0.96$, respectively. Note that a higher ``pred\_iou\_thresh" value tends to generate more dependable raw output. The threshold $K$ and $S$ are both set to $50$. Stochastic gradient descent (SGD) was employed as the optimization algorithm for training all models. Furthermore, the experiments employed a learning rate of $0.01$, a momentum of $0.9$, a decaying coefficient of $0.0005$, and a batch size of $10$. The values of $\lambda_{o}$ and $\lambda_{b}$ were chosen based on our sensitivity analysis as shown in Section~\ref{sec:sen}. Simple data augmentations, such as random rotation and flipping, were utilized for all experiments.

\begin{table*}\footnotesize
  \centering
	\caption{Experimental results on the LoveDA Urban dataset. We present the OA of five foreground classes and three overall performance metrics.  The accuracy of each category is presented in the F1/IoU form. Bold values are the best.}
	  \tabcolsep=0.12cm
	  \renewcommand\arraystretch{1.2}
		\begin{tabular}{ccccccccccc}
			\hline
			\textbf{Method}  & \textbf{Backbone}  & \textbf{background} & \textbf{building} & \textbf{road}  & \textbf{water}  & \textbf{barren}  & \textbf{forest} & \textbf{agriculture}  & \textbf{mF1} & \textbf{mIoU}  \\
			\hline
			ABCNet  \cite{li2021abcnet}  &ResNet-18& \textbf{52.02/35.15} & 63.36/46.37 & 65.42/48.61 & 61.42/44.31 & \textbf{44.27/28.43} & 54.63/37.58 & 19.98/11.10 & 57.30 & 40.58 \\
			ABCNet+SAM  &ResNet-18& 50.23/33.54 & \textbf{68.25/51.80} & \textbf{66.10/49.36} & \textbf{75.74/60.95} & 36.09/22.01 & \textbf{55.43/38.34} & \textbf{22.59/12.74} & \textbf{59.28} & \textbf{43.53} \\
			\hline
			CMTFNet \cite{wu2023cmtfnet} &ResNet-50& \textbf{56.09/38.98} & 74.18/58.96 & 67.11/50.50 & 70.3/554.27 & \textbf{47.00/30.72} & 54.45/37.41 & 41.92/25.65 & 62.95 & 46.68 \\
			CMTFNet+SAM  &ResNet-50& 55.86/38.76 & \textbf{75.79/61.02} & \textbf{76.05/61.36} & \textbf{72.14/56.42} & 37.28/22.91 & \textbf{55.97/38.86} & \textbf{43.24/27.58} & \textbf{63.43} & \textbf{48.09} \\
			\hline
			UNetformer  \cite{wang2022unetformer} &ResNet-18& 54.66/37.61 & 69.09/52.78 & 68.33/51.89 & 77.66/63.47 & \textbf{56.98/39.84} & 51.01/34.23 & 20.54/11.44 & 65.34 & 49.12 \\
		  UNetformer+SAM  &ResNet-18& \textbf{55.43/38.34} & \textbf{75.91/61.17} & \textbf{73.61/58.24} & \textbf{80.55/67.43} & 43.19/27.55 & \textbf{57.05/39.91} & \textbf{43.64/27.91} & \textbf{65.74} & \textbf{50.55} \\
			\hline	
			FTUNetformer \cite{wang2022unetformer} &SwinTrans-base& 55.38/38.30 & 75.46/61.21 & \textbf{75.29/60.38} & 75.81/61.04 & 40.83/25.65 & \textbf{54.10/37.08} & 40.32/25.25 & 64.95 & 49.71 \\
			FTUNetformer+SAM &SwinTrans-base& \textbf{56.40/39.27} & \textbf{77.23/62.91} & 74.74/59.67 & \textbf{77.30/63.00} & \textbf{44.44/28.57} & 53.30/36.33 & \textbf{44.17/28.34} & \textbf{66.02} & \textbf{50.68} \\
			\hline
	\end{tabular}\label{tab:ulist}
\end{table*}

\subsection{Performance Comparison}
We benchmarked the performance of the proposed framework on four representative semantic segmentation models for remote sensing images, namely ABCNet \cite{li2021abcnet}, CMTFNet \cite{wu2023cmtfnet}, UNetformer \cite{wang2022unetformer} and FTUNetformer \cite{wang2022unetformer}. Specifically, these four methods employ the most common backbones recently, the CNN-based ResNet \cite{he2016deep} and the self-attention-based \cite{vaswani2017attention} Swin Transformer \cite{liu2021swin}. This choice allows us to validate our approach across different classical networks. The quantitative results are listed in Table~\ref{tab:vlist} and Table~\ref{tab:ulist}.

\subsubsection{Performance Comparison on the Vaihingen dataset}
As indicated in Table~\ref{tab:vlist}, our proposed framework demonstrates significant improvements in terms of both mF1 and mIoU metrics as compared to the four baseline methods. These results affirm the effectiveness of our approach in robustly capturing object and boundary representations. Notably, the incorporation of SGO and SGB information substantially enhances semantic segmentation performance in remote sensing images. For illustration purposes, we focus on the third set of experiments in which UNetformer+SAM demonstrated superior performance over UNetformer in all five classes, namely {\em impervious surface}, {\em building}, {\em low vegetation}, {\em tree}, and {\em car}. In particular, on the ISPRS Vaihingen dataset, UNetformer+SAM achieved performance improvement of $1.08\%$ and $1.78\%$ in F1 and IoU, respectively, on the {\em Car} class as compared to UNetformer. Furthermore, the classification accuracy for {\em building} was enhanced by $0.49\%$ and $0.91\%$ in F1 and IoU, respectively, compared to UNetformer. These improvements can be attributed to the inherent characteristics of {\em car} and {\em building} classes, which typically exhibit simpler and more standardized shapes compared to other ground object categories. For instance, a car is often presented as a fixed-size rectangular shape under consistent sampling conditions, while a building typically manifests as a changing-size rectangle. As a result, SGO and SGB provide more reliable criteria for these specific categories, facilitating more accurate segmentation.

In contrast, {\em impervious surface}, {\em low vegetation}, and {\em tree} often feature intricate boundaries and varying sizes. Despite their complexity, SGO and SGB still offer valuable supplementary information for these specific categories. For instance, the classification accuracy for {\em impervious surface} was improved by $0.46\%$ and $0.78\%$ in F1 and IoU, respectively. Meanwhile, both {\em low vegetation} and {\em tree} showed comparable improvements in F1 and IoU metrics, reflecting their similar characteristics. In terms of overall performance, UNetformer+SAM achieved an mF1 score of $90.40\%$ and a mIoU of $82.88\%$, representing an increase of $0.55\%$ and $0.91\%$ compared to UNetformer, respectively.

Other comparative experiments in Table~\ref{tab:vlist} further substantiated our observations above. Specifically, FTUNetformer+SAM that adopts Swin Transformer as its backbone, also demonstrated significant enhancements, exhibiting a notable $2.63\%$ and $4.33\%$ improvement in F1 and IoU metrics, respectively, for the {\em Car} class compared to FTUNetformer. Moreover, we observed a slight increase in classification accuracy for the {\em building}, with improvements of $0.27\%$ and $0.51\%$ in F1 and IoU, respectively. In particular, the {\em tree} category exhibits a notable improvement, while the improvement in the {\em low vegetation} category is somewhat diminished, primarily due to their significant similarity. The overall performance corresponds to an mF1 score of $91.08\%$ and an mIoU of $84.01\%$, marking respective increases of $0.84\%$ and $1.39\%$ compared to the FTUNetformer. These outcomes affirm that our proposed framework, with the aid of SGO and SGB, achieved superior generalization performance across various models.

\begin{figure*}[hbp]
\centering
{\includegraphics[width=0.85\linewidth]{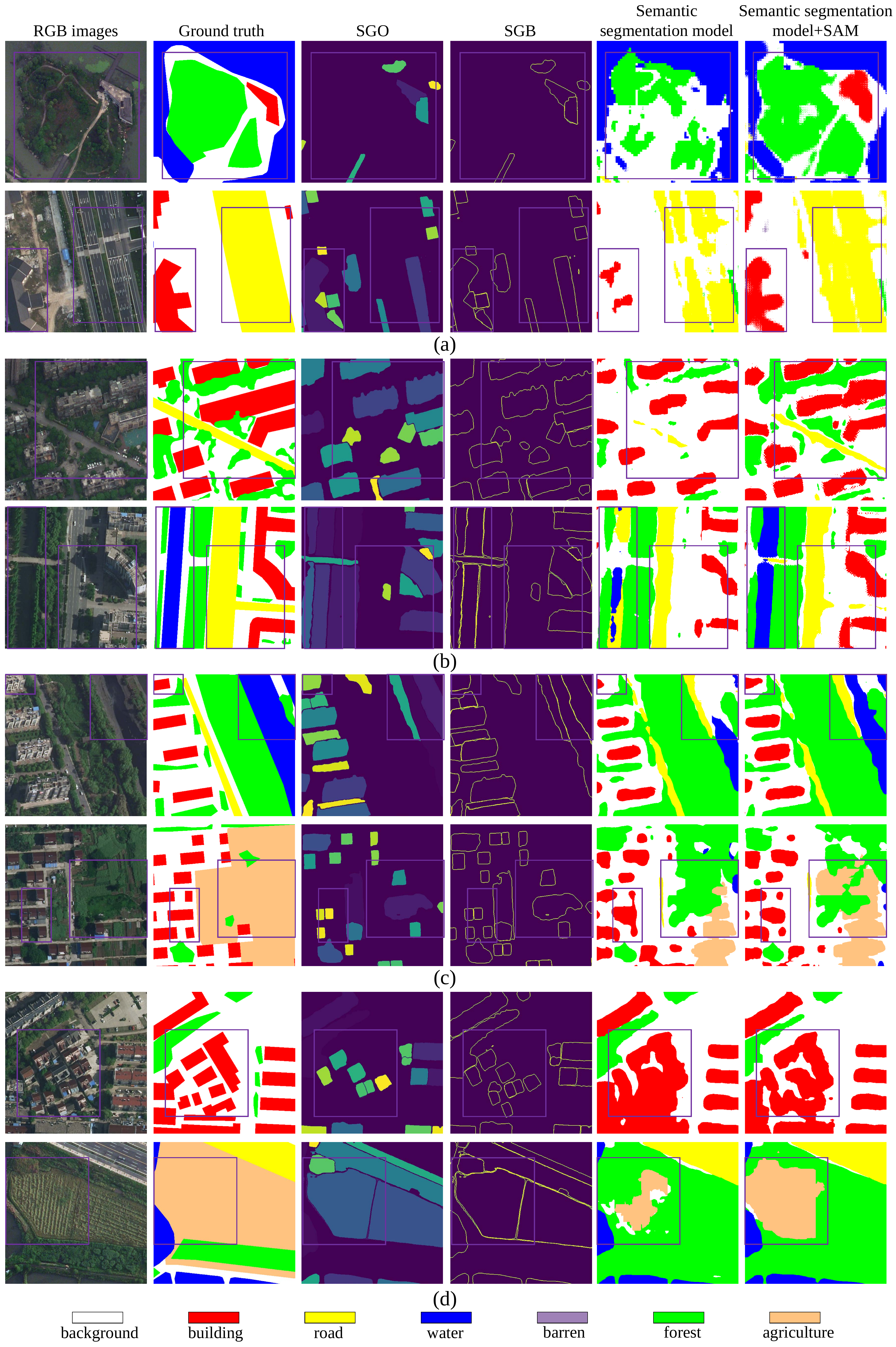}}
\caption{Qualitative performance comparisons on the LoveDA Urban with the size of $512 \times 512$. (a) ABCNet, (b) CMTFNet, (c) UNetformer, (d) FTUNetformer. We showcase two samples for each model. Some purple boxes are added to highlight the differences.}
\label{fig4}
\end{figure*}

Fig. \ref{fig3} presents a visual comparison between the results produced by the baseline methods and our approach. Throughout all subfigures, purple boxes highlight areas of interest. Firstly, our method excelled in precisely segmenting entire objects, notably evident in the clear delineation of {\em building}. Secondly, our approach significantly refined object boundaries, as seen {\em low vegetation} and {\em building}, aligning more closely with the actual ground truth boundaries. Furthermore, our method successfully addressed the segmentation of challenging categories, such as {\em tree}, {\em building} and {\em impervious surface}. These improvements primarily stem from the object region and boundary details contained in the SGO and SGB. Leveraging the well-designed framework founded on object- and boundary-based loss functions allows for the comprehensive exploitation of these detailed information, leading to the observed improvements in segmentation accuracy.

\subsubsection{Performance Comparison on the LoveDA Urban}
Experiments conducted on the LoveDA Urban dataset yielded results similar to those observed in the ISPRS Vaihingen dataset despite variations in sampling resolution and ground object categories between the two datasets. Given the modest benchmark performance of ABCNet and CMTFNet, our approach yielded notably superior enhancements in overall performance. Observing the results of UNetformer and UNetformer+SAM as shown in Table~\ref{tab:ulist}, the classification accuracy for {\em building}, {\em water}, and {\em agriculture} were $75.91\%/61.17\%$, $80.55\%/67.43\%$, and $43.64\%/27.91\%$, respectively, which amounts to an increase of $6.82\%/8.39\%$, $2.89\%/3.96\%$, and $23.1\%/16.47\%$ in F1 and IoU, respectively, as compared to UNetformer. Meanwhile, our findings with FTUNetformer+SAM consistently confirmed advantages for these classes characterized by regular shapes and uncomplicated borders. The corresponding mF1 and mIoU values were $65.74\%$, $50.55\%$ respectively, marking increases of $0.4\%$, and $1.43\%$, respectively, over UNetformer while the increases for mF1 and mIoU on FTUNetformer were $1.07\%$, $0.97\%$. 

However, upon evaluating this dataset, we observed fluctuating performance changes when comparing the results of the four baselines. For instance, UNetformer+SAM exhibited improved performance for {\em road} and {\em forest}, whereas FTUNetformer+SAM demonstrated the opposite trend. Conversely, the performance for {\em barren} decreased in UNetformer+SAM but improved in FTUNetformer+SAM. The primary reason for this discrepancy is that these three categories have intricate boundaries and varying sizes. Notably, the proportion of {\em barren} is very small, leading to volatility in its results during the experiments \cite{wang2021loveda}. While the semantic segmentation model aims to enhance overall performance, it cannot guarantee consistent improvement in every single category. Nevertheless, the overall performance improvement of our framework demonstrates the instructive nature of this approach for subsequent remote sensing works with different models.

Fig. \ref{fig4} presents visualization examples from the LoveDA Urban, with purple boxes highlighting areas of interest in all subfigures. Firstly, the model demonstrated enhanced accuracy in identifying {\em building}, evident across all methods. Furthermore, a more complete identification of {\em water} and {\em agriculture} objects was apparent. These observations closely align with the improvements indicated by the mF1 and mIoU metrics as presented in Table~\ref{tab:ulist}.

\begin{figure}[t]
\centering
{\includegraphics[width=0.95\linewidth]{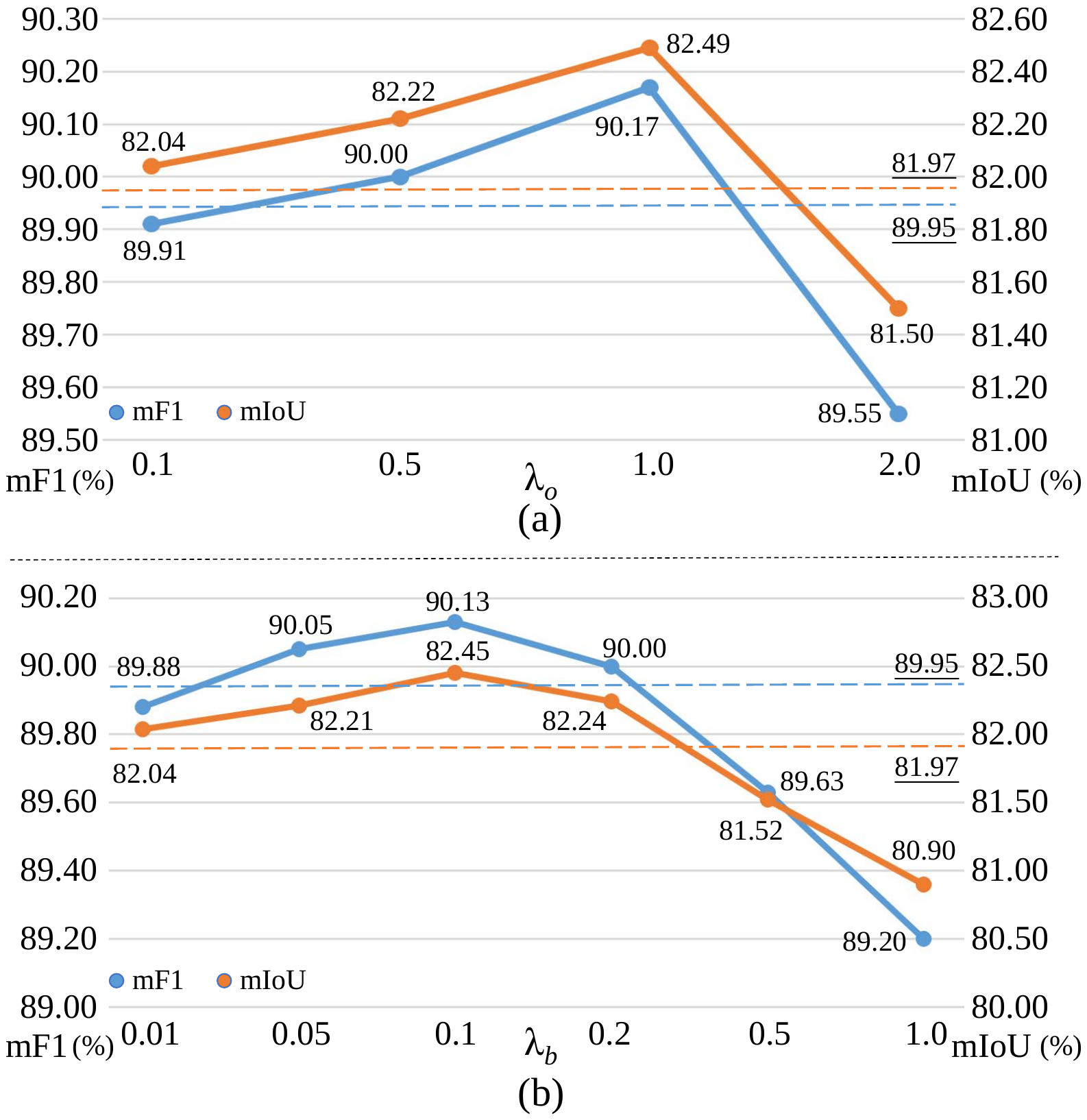}}
\caption{Sensitivity analysis of (a) $\lambda_{o}$ and (b) $\lambda_{b}$ on the ISPRS Vaihingen dataset. The performance of the baseline is depicted using dashed lines to highlight the variation of performance.}
\label{fig4}
\end{figure}

\begin{table*}[t]\footnotesize
	\centering
	\caption{Ablation studies on the ISPRS Vaihingen dataset. The accuracy of each category is presented in the F1/IoU form. Bold values are the best.}
		\renewcommand\arraystretch{1.5}
		\begin{tabular}{ccccccccc}
			\hline
			\textbf{Method}    & \textbf{impervious surface} & \textbf{building} & \textbf{low vegetation}  & \textbf{tree}  & \textbf{car}  & \textbf{mF1}    & \textbf{mIoU}  \\
			\hline
			$L_{seg}$             & 92.33/85.76 & 96.25/92.78 & 80.47/67.33 & 90.85/83.22 & 89.35/80.75 & 89.85 & 81.97  \\
                $L_{seg}$ + $L_{obj}$    & 92.55/86.14 & 96.48/93.20 & 80.76/67.73 & 91.14/83.72 & 89.91/81.67 & 90.17 & 82.49  \\
                $L_{seg}$ + $L_{bdy}$  & \textbf{92.79}/\textbf{86.54} & \textbf{96.79}/\textbf{93.78} & 80.66/67.59 & 90.94/83.38 & 89.49/80.99 & 90.13 & 82.45  \\
                $L_{seg}$ + $L_{obj}$ + $L_{bdy}$  & \textbf{92.79}/\textbf{86.54} & 96.74/93.69 & \textbf{80.85/67.86} & \textbf{91.17/83.77} & \textbf{90.43/82.5}3 & \textbf{90.40} & \textbf{82.88}  \\
			\hline
	\end{tabular}\label{tab:vabla}
\end{table*}

\begin{table*}[t]\footnotesize
	\centering
	\caption{Ablation studies on the LoveDA Urban dataset. The accuracy of each category is presented in the F1/IoU form. Bold values are the best.}
		\renewcommand\arraystretch{1.5}
		\begin{tabular}{cccccccccc}
			\hline
			\textbf{Method}    & \textbf{background} & \textbf{building} & \textbf{road}  & \textbf{water}  & \textbf{barren}  & \textbf{forest}  & \textbf{agriculture}  & \textbf{mF1}    & \textbf{mIoU}  \\
			\hline
			$L_{seg}$             & 54.66/37.61 & 69.09/52.78 & 68.33/51.89 & 77.66/63.47 & \textbf{56.98/39.84} & 51.01/34.23 & 20.54/11.44 & 65.34 & 49.12 \\
                $L_{seg}$ + $L_{obj}$    & 53.04/36.08 & 72.54/56.91 & 71.68/55.86 & 75.62/60.79 & 54.76/37.70 & \textbf{61.00/43.89} & 23.50/13.32 & 65.53 & 49.47 \\
                $L_{seg}$ + $L_{bdy}$  & \textbf{55.49/38.39} & 72.73/57.15 & 73.30/57.86 & 79.33/65.74 & 48.54/32.05 & 54.54/37.49 & 41.13/25.89 & \textbf{65.88} & 50.24  \\
                $L_{seg}$ + $L_{obj}$ + $L_{bdy}$  & 55.43/38.34 & \textbf{75.91/61.17} & \textbf{73.61/58.24} & \textbf{80.55/67.43} & 43.19/27.55 & 57.05/39.91 & \textbf{43.64/27.91} & 65.74 & \textbf{50.55} \\
			\hline
	\end{tabular}\label{tab:uabla}
\end{table*}

\begin{table*}[t]
	\centering
	\caption{Computational complexity analysis measured by two $256 \times 256$ images on a single NVIDIA GeForce RTX 4090 GPU. MIoU values are the results on the ISPRS Vaihingen dataset. Bold values are the best.}
	\setlength{\tabcolsep}{5mm}{
		\begin{tabular}{m{2.3cm}<{\centering}|m{1.2cm}<{\centering}m{1.2cm}<{\centering}m{1.2cm}<{\centering}m{1.2cm}<{\centering}m{1.2cm}<{\centering}m{1cm}<{\centering}}
			\hline
			\textbf{Model}   & \textbf{FLOPs (G)}  &  \textbf{Parameter (M)} & \textbf{Memory (MB)}  &  \textbf{Running Time ($10^{-3}$s)} &  \textbf{Inference Speed (FPS)} &  \textbf{MIoU(\%)}  \\
			\hline
			ABCNet \cite{li2021abcnet} & \textbf{7.81} & \textbf{13.67} & \textbf{2480} & \textbf{1.23} & \textbf{32.62} & 75.20 \\
			ABCNet+SAM                 & \textbf{7.81} & \textbf{13.67} & \textbf{2480} & 4.99 & \textbf{32.62} & \textbf{78.11} \\
			\hline
			CMTFNet \cite{wu2023cmtfnet}   & \textbf{17.14} & \textbf{30.07} & \textbf{3240} & \textbf{1.84} & \textbf{14.47} & 82.12 \\
			CMTFNet+SAM                    & \textbf{17.14} & \textbf{30.07} & \textbf{3240} & 5.97 & \textbf{14.47} & \textbf{83.18} \\
			\hline
			UNetformer \cite{wang2022unetformer} & \textbf{5.87} & \textbf{11.69} & \textbf{2490} & \textbf{14.35} & \textbf{31.11} & 81.97 \\
			UNetformer+SAM                       & \textbf{5.87} & \textbf{11.69} & \textbf{2490} & 50.56 & \textbf{31.11} & \textbf{82.88} \\
			\hline
			FTUNetformer \cite{wang2022unetformer}  & \textbf{50.84} & \textbf{96.14} & \textbf{5176} & \textbf{3.40} & \textbf{21.35} & 82.62 \\
			FTUNetformer+SAM                        & \textbf{50.84} & \textbf{96.14} & \textbf{5176} & 74.85 & \textbf{21.35} & \textbf{84.01} \\
			\hline
	\end{tabular}}\label{tab:scale}
\end{table*}

\subsection{Sensitivity Analysis}\label{sec:sen}
The analysis focuses on the sensitivity of two hyper-parameters, $\lambda_{o}$ and $\lambda_{b}$. $\lambda_{o}$ is designed to adjust the influence stemming from object consistency, whereas $\lambda_{b}$ adjusts the contribution originating from boundary information. These parameters hold significance in balancing the object consistency loss and boundary preservation loss. Considering that there is no strong correlation between these hyper-parameters, separate sensitivity experiments were conducted using UNetformer.

Fig.~\ref{fig4} (a) illustrates the model performance in terms of mF1 and mIoU across varying $\lambda_{o}$ values. A smaller $\lambda_{o}$ diminishes the influence of object information, while a larger value can overly emphasize its significance. Notably, setting $\lambda_{o}\geq 2.0$ resulted in a noticeable degradation in performance. Conversely, within the range $\lambda_{o}\in \left[0.1,1.0\right]$, performance exhibited lower sensitivity to changes in $\lambda_{o}$. Hereby our experiments employed $\lambda_{o}=1.0$, unless specified otherwise. Additionally, in Fig.~\ref{fig4} (b), the model's performance is depicted concerning varying $\lambda_{b}\in \left[0.01,1.0\right]$. A smaller $\lambda_{b}$ value made the contribution from the boundary information insignificant in learning the model. The best performance was observed when $\lambda_{b}$ reached $0.1$, as illustrated in Fig.~\ref{fig4} (b). However, further increments in $\lambda_{b}$ resulted in performance degradation. Ultimately, our experiments set $\lambda_{o}$ and $\lambda_{b}$ to $1.0$ and $0.1$, respectively, aligning them within a similar order of magnitude.

\subsection{Ablation Study}
To emphasize the distinct roles of the two loss functions in the proposed framework, we conducted ablation experiments using UNetformer on two datasets, as detailed in Table~\ref{tab:vabla} and Table~\ref{tab:uabla}. The results underscore that the independent use of these two loss functions enhanced the overall performance of the semantic segmentation model, validating the value and efficacy of SAM's raw output. When examining the results on {\em impervious surface} and {\em building} of the ISPRS Vaihingen dataset, it is evident that only the boundary preservation loss exhibited comparable improvement as compared to the combined loss. This suggests that SAM-based pre-processing results can be fully explored using only the boundary preservation loss in specific tasks, such as building detection. However, in semantic segmentation tasks, where various ground objects possess highly complex boundaries, leveraging both SGO and SGB to their fullest extent becomes necessary. Furthermore, the results of these two loss functions on different categories, as listed in Table~\ref{tab:vabla} and Table~\ref{tab:uabla}, highlight the disparate behaviors exhibited across various categories. Combining the object and boundary preservation loss functions, we proposed a versatile and streamlined framework for directly leveraging SGO and SGB, showcasing robust generalization capabilities in remote sensing image semantic segmentation tasks.

\subsection{Model Complexity Analysis}\label{sec:complexity}
We evaluate the computational complexity of the proposed framework using various metrics: floating point operation count (FLOPs), model parameter, memory footprint, running time (s) and frames per second (FPS). FLOPs accesses the model's complexity, while model parameters and memory footprint evaluate the scale of the network and memory requirements, respectively. Finally, running time and FPS quantifies and training speed and inference speed, respectively. Ideally, an efficient model maintains lower values in FLOPs, model parameters, memory footprint and running time while achieving a higher FPS value.

Table~\ref{tab:scale} shows the complexity analysis results of all comparing semantic segmentation models considered in this work. Inspection of Table~\ref{tab:scale} shows that our approach introduces no extra model complexity or inference time. In our framework, it is necessary to generate SGO and SGB before their utilization in loss calculation, hereby negating the requirement for additional task-specific modules. However, it is observed that our approach lengthens the training time since the gradient back-propagation takes longer to compute with two additional loss functions. On the other hand, during inference stage as presented in Fig.~\ref{fig2} (b), the proposed approach operates identically to the original model, ensuring zero impact on inference speed. Considering the enhancements observed in other aspects, we believe that the marginal rise in training time is a justifiable trade-off. These outcomes highlight the remarkable scalability and wide application of our framework across current semantic segmentation models.

\section{Conclusion}\label{sec:con}
This work proposed a simple and versatile framework designed to fully exploit the raw output of SAM in conjunction with general remote sensing imagery semantic segmentation models. Acknowledging the distinctions between remote sensing images and natural images, as well as the characteristics of SGO and SGB, we developed an auxiliary optimization strategy by exploiting two loss functions, object consistency loss and boundary preservation loss. This strategy facilitates improvements in fundamental semantic segmentation tasks with different network structures without requiring additional modules. Notably, our introduction of object consistency loss, motivated by the consistency of objects, represents the initial loss function capable of directly utilizing SGO without semantic information. Our validation on two publicly available datasets and four comprehensive semantic segmentation models highlights the robust performance of our framework. Finally, it is worth emphasizing that this work has initiated a preliminary exploration of SAM's raw output, revealing the potential for large models such as SAM in remote sensing. 

\small
\bibliographystyle{IEEEtran}
\bibliography{references}

\begin{thebibliography}{10}
\providecommand{\url}[1]{#1}
\csname url@samestyle\endcsname
\providecommand{\newblock}{\relax}
\providecommand{\bibinfo}[2]{#2}
\providecommand{\BIBentrySTDinterwordspacing}{\spaceskip=0pt\relax}
\providecommand{\BIBentryALTinterwordstretchfactor}{4}
\providecommand{\BIBentryALTinterwordspacing}{\spaceskip=\fontdimen2\font plus
\BIBentryALTinterwordstretchfactor\fontdimen3\font minus
  \fontdimen4\font\relax}
\providecommand{\BIBforeignlanguage}[2]{{%
\expandafter\ifx\csname l@#1\endcsname\relax
\typeout{** WARNING: IEEEtran.bst: No hyphenation pattern has been}%
\typeout{** loaded for the language `#1'. Using the pattern for}%
\typeout{** the default language instead.}%
\else
\language=\csname l@#1\endcsname
\fi
#2}}
\providecommand{\BIBdecl}{\relax}
\BIBdecl

\bibitem{yuan2020deep}
Q.~Yuan, H.~Shen, T.~Li, Z.~Li, S.~Li, Y.~Jiang, H.~Xu, W.~Tan, Q.~Yang,
  J.~Wang \emph{et~al.}, ``Deep learning in environmental remote sensing:
  Achievements and challenges,'' \emph{Remote Sensing of Environment}, vol.
  241, p. 111716, 2020.

\bibitem{cao2022coarse}
Y.~Cao and X.~Huang, ``A coarse-to-fine weakly supervised learning method for
  green plastic cover segmentation using high-resolution remote sensing
  images,'' \emph{ISPRS Journal of Photogrammetry and Remote Sensing}, vol.
  188, pp. 157--176, 2022.

\bibitem{li2022land}
R.~Li, S.~Zheng, C.~Duan, L.~Wang, and C.~Zhang, ``Land cover classification
  from remote sensing images based on multi-scale fully convolutional
  network,'' \emph{Geo-spatial information science}, vol.~25, no.~2, pp.
  278--294, 2022.

\bibitem{xu2023rssformer}
R.~Xu, C.~Wang, J.~Zhang, S.~Xu, W.~Meng, and X.~Zhang, ``Rssformer: Foreground
  saliency enhancement for remote sensing land-cover segmentation,'' \emph{IEEE
  Transactions on Image Processing}, vol.~32, pp. 1052--1064, 2023.

\bibitem{gupta2021deep}
A.~Gupta, S.~Watson, and H.~Yin, ``Deep learning-based aerial image
  segmentation with open data for disaster impact assessment,''
  \emph{Neurocomputing}, vol. 439, pp. 22--33, 2021.

\bibitem{khan2021deepsmoke}
S.~Khan, K.~Muhammad, T.~Hussain, J.~Del~Ser, F.~Cuzzolin, S.~Bhattacharyya,
  Z.~Akhtar, and V.~H.~C. de~Albuquerque, ``Deepsmoke: Deep learning model for
  smoke detection and segmentation in outdoor environments,'' \emph{Expert
  Systems with Applications}, vol. 182, p. 115125, 2021.

\bibitem{huang2022evaluation}
D.~Huang, Y.~Tang, and R.~Qin, ``An evaluation of planetscope images for 3d
  reconstruction and change detection--experimental validations with case
  studies,'' \emph{GIScience \& Remote Sensing}, vol.~59, no.~1, pp. 744--761,
  2022.

\bibitem{bo2022basnet}
W.~Bo, J.~Liu, X.~Fan, T.~Tjahjadi, Q.~Ye, and L.~Fu, ``Basnet: Burned area
  segmentation network for real-time detection of damage maps in remote sensing
  images,'' \emph{IEEE Transactions on Geoscience and Remote Sensing}, vol.~60,
  pp. 1--13, 2022.

\bibitem{ronneberger2015u}
O.~Ronneberger, P.~Fischer, and T.~Brox, ``{U-Net}: Convolutional networks for
  biomedical image segmentation,'' in \emph{Medical Image Computing and
  Computer-Assisted Intervention--MICCAI 2015: 18th International Conference,
  Munich, Germany, October 5-9, 2015, Proceedings, Part III 18}.\hskip 1em plus
  0.5em minus 0.4em\relax Springer, 2015, pp. 234--241.

\bibitem{he2016deep}
K.~He, X.~Zhang, S.~Ren, and J.~Sun, ``Deep residual learning for image
  recognition,'' in \emph{Proceedings of the IEEE conference on computer vision
  and pattern recognition}, 2016, pp. 770--778.

\bibitem{vaswani2017attention}
A.~Vaswani, N.~Shazeer, N.~Parmar, J.~Uszkoreit, L.~Jones, A.~N. Gomez,
  {\L}.~Kaiser, and I.~Polosukhin, ``Attention is all you need,''
  \emph{Advances in neural information processing systems}, vol.~30, 2017.

\bibitem{dosovitskiy2020image}
A.~Dosovitskiy, L.~Beyer, A.~Kolesnikov, D.~Weissenborn, X.~Zhai,
  T.~Unterthiner, M.~Dehghani, M.~Minderer, G.~Heigold, S.~Gelly \emph{et~al.},
  ``An image is worth 16x16 words: Transformers for image recognition at
  scale,'' \emph{arXiv preprint arXiv:2010.11929}, 2020.

\bibitem{diakogiannis2020resunet}
F.~I. Diakogiannis, F.~Waldner, P.~Caccetta, and C.~Wu, ``{ResUNet-a}: A deep
  learning framework for semantic segmentation of remotely sensed data,''
  \emph{ISPRS Journal of Photogrammetry and Remote Sensing}, vol. 162, pp.
  94--114, 2020.

\bibitem{FuseNet}
C.~Hazirbas, L.~Ma, C.~Domokos, and D.~Cremers, ``{FuseNet}: Incorporating
  depth into semantic segmentation via fusion-based {CNN} architecture,'' in
  \emph{Asian conference on computer vision}, 2016, pp. 213--228.

\bibitem{hong2020multimodal}
D.~Hong, J.~Yao, D.~Meng, Z.~Xu, and J.~Chanussot, ``Multimodal {GANs}: Toward
  crossmodal hyperspectral--multispectral image segmentation,'' \emph{IEEE
  Transactions on Geoscience and Remote Sensing}, vol.~59, no.~6, pp.
  5103--5113, 2020.

\bibitem{hong2023cross}
D.~Hong, B.~Zhang, H.~Li, Y.~Li, J.~Yao, C.~Li, M.~Werner, J.~Chanussot,
  A.~Zipf, and X.~X. Zhu, ``Cross-city matters: A multimodal remote sensing
  benchmark dataset for cross-city semantic segmentation using high-resolution
  domain adaptation networks,'' \emph{Remote Sensing of Environment}, vol. 299,
  p. 113856, 2023.

\bibitem{wang2022novel}
L.~Wang, R.~Li, C.~Duan, C.~Zhang, X.~Meng, and S.~Fang, ``A novel transformer
  based semantic segmentation scheme for fine-resolution remote sensing
  images,'' \emph{IEEE Geoscience and Remote Sensing Letters}, vol.~19, pp.
  1--5, 2022.

\bibitem{xu2021efficient}
Z.~Xu, W.~Zhang, T.~Zhang, Z.~Yang, and J.~Li, ``Efficient transformer for
  remote sensing image segmentation,'' \emph{Remote Sensing}, vol.~13, no.~18,
  p. 3585, 2021.

\bibitem{roy2023multimodal}
S.~K. Roy, A.~Deria, D.~Hong, B.~Rasti, A.~Plaza, and J.~Chanussot,
  ``Multimodal fusion transformer for remote sensing image classification,''
  \emph{IEEE Transactions on Geoscience and Remote Sensing}, 2023.

\bibitem{yang2023gtfn}
A.~Yang, M.~Li, Y.~Ding, D.~Hong, Y.~Lv, and Y.~He, ``{GTFN}: {GCN} and
  transformer fusion with spatial-spectral features for hyperspectral image
  classification,'' \emph{IEEE Transactions on Geoscience and Remote Sensing},
  2023.

\bibitem{wang2022unetformer}
L.~Wang, R.~Li, C.~Zhang, S.~Fang, C.~Duan, X.~Meng, and P.~M. Atkinson,
  ``{UNetFormer}: A {UNet}-like transformer for efficient semantic segmentation
  of remote sensing urban scene imagery,'' \emph{ISPRS Journal of
  Photogrammetry and Remote Sensing}, vol. 190, pp. 196--214, 2022.

\bibitem{zhang2022transformer}
C.~Zhang, W.~Jiang, Y.~Zhang, W.~Wang, Q.~Zhao, and C.~Wang, ``Transformer and
  {CNN} hybrid deep neural network for semantic segmentation of
  very-high-resolution remote sensing imagery,'' \emph{IEEE Transactions on
  Geoscience and Remote Sensing}, vol.~60, pp. 1--20, 2022.

\bibitem{ma2023unsupervised}
X.~Ma, X.~Zhang, Z.~Wang, and M.-O. Pun, ``Unsupervised domain adaptation
  augmented by mutually boosted attention for semantic segmentation of {VHR}
  remote sensing images,'' \emph{IEEE Transactions on Geoscience and Remote
  Sensing}, vol.~61, pp. 1--15, 2023.

\bibitem{wu2023cmtfnet}
H.~Wu, P.~Huang, M.~Zhang, W.~Tang, and X.~Yu, ``{CMTFNet}: {CNN} and
  multiscale transformer fusion network for remote sensing image semantic
  segmentation,'' \emph{IEEE Transactions on Geoscience and Remote Sensing},
  2023.

\bibitem{kirillov2023segment}
A.~Kirillov, E.~Mintun, N.~Ravi, H.~Mao, C.~Rolland, L.~Gustafson, T.~Xiao,
  S.~Whitehead, A.~C. Berg, W.-Y. Lo \emph{et~al.}, ``Segment anything,''
  \emph{arXiv preprint arXiv:2304.02643}, 2023.

\bibitem{ren2023segment}
S.~Ren, F.~Luzi, S.~Lahrichi, K.~Kassaw, L.~M. Collins, K.~Bradbury, and J.~M.
  Malof, ``Segment anything, from space?'' \emph{arXiv preprint
  arXiv:2304.13000}, 2023.

\bibitem{stearns2023segment}
L.~Stearns, C.~van~der Veen, and S.~Shankar, ``Segment anything in glaciology:
  An initial study implementing the segment anything model ({SAM}),'' 2023.

\bibitem{ding2023adapting}
L.~Ding, K.~Zhu, D.~Peng, H.~Tang, and H.~Guo, ``Adapting segment anything
  model for change detection in {HR} remote sensing images,'' \emph{arXiv
  preprint arXiv:2309.01429}, 2023.

\bibitem{zhang2023enhancing}
C.~Zhang, P.~Marfatia, H.~Farhan, L.~Di, L.~Lin, H.~Zhao, H.~Li, M.~D. Islam,
  and Z.~Yang, ``Enhancing {USDA} {NASS} cropland data layer with segment
  anything model,'' in \emph{2023 11th International Conference on
  Agro-Geoinformatics (Agro-Geoinformatics)}.\hskip 1em plus 0.5em minus
  0.4em\relax IEEE, 2023, pp. 1--5.

\bibitem{osco2023segment}
L.~P. Osco, Q.~Wu, E.~L. de~Lemos, W.~N. Gon{\c{c}}alves, A.~P.~M. Ramos,
  J.~Li, and J.~M. Junior, ``The segment anything model ({SAM}) for remote
  sensing applications: From zero to one shot,'' \emph{International Journal of
  Applied Earth Observation and Geoinformation}, vol. 124, p. 103540, 2023.

\bibitem{li2023rs}
X.~Li, C.~Wen, Y.~Hu, and N.~Zhou, ``{RS-CLIP}: Zero shot remote sensing scene
  classification via contrastive vision-language supervision,''
  \emph{International Journal of Applied Earth Observation and Geoinformation},
  vol. 124, p. 103497, 2023.

\bibitem{qi2023self}
X.~Qi, Y.~Wu, Y.~Mao, W.~Zhang, and Y.~Zhang, ``Self-guided few-shot semantic
  segmentation for remote sensing imagery based on large vision models,''
  \emph{arXiv preprint arXiv:2311.13200}, 2023.

\bibitem{al2023vision}
M.~M. Al~Rahhal, Y.~Bazi, H.~Elgibreen, and M.~Zuair, ``Vision-language models
  for zero-shot classification of remote sensing images,'' \emph{Applied
  Sciences}, vol.~13, no.~22, p. 12462, 2023.

\bibitem{chen2023rsprompter}
K.~Chen, C.~Liu, H.~Chen, H.~Zhang, W.~Li, Z.~Zou, and Z.~Shi, ``{RSPrompter}:
  Learning to prompt for remote sensing instance segmentation based on visual
  foundation model,'' \emph{arXiv preprint arXiv:2306.16269}, 2023.

\bibitem{SAMRS}
D.~Wang, J.~Zhang, B.~Du, M.~Xu, L.~Liu, D.~Tao, and L.~Zhang, ``{SAMRS}:
  Scaling-up remote sensing segmentation dataset with segment anything model,''
  in \emph{Thirty-seventh Conference on Neural Information Processing Systems
  Datasets and Benchmarks Track}, 2023.

\bibitem{zhang2023text2seg}
J.~Zhang, Z.~Zhou, G.~Mai, L.~Mu, M.~Hu, and S.~Li, ``{Text2Seg}: Remote
  sensing image semantic segmentation via text-guided visual foundation
  models,'' \emph{arXiv preprint arXiv:2304.10597}, 2023.

\bibitem{sultan2023geosam}
R.~I. Sultan, C.~Li, H.~Zhu, P.~Khanduri, M.~Brocanelli, and D.~Zhu,
  ``{GeoSAM}: Fine-tuning {SAM} with sparse and dense visual prompting for
  automated segmentation of mobility infrastructure,'' \emph{arXiv preprint
  arXiv:2311.11319}, 2023.

\bibitem{julka2023knowledge}
S.~Julka and M.~Granitzer, ``Knowledge distillation with segment anything
  ({SAM}) model for planetary geological mapping,'' \emph{arXiv preprint
  arXiv:2305.07586}, 2023.

\bibitem{yan2023ringmo}
Z.~Yan, J.~Li, X.~Li, R.~Zhou, W.~Zhang, Y.~Feng, W.~Diao, K.~Fu, and X.~Sun,
  ``Ringmo-sam: A foundation model for segment anything in multimodal
  remote-sensing images,'' \emph{IEEE Transactions on Geoscience and Remote
  Sensing}, vol.~61, pp. 1--16, 2023.

\bibitem{wang2023cs}
L.~Wang, M.~Zhang, and W.~Shi, ``{CS-WSCDNet}: Class activation mapping and
  segment anything model-based framework for weakly supervised change
  detection,'' \emph{IEEE Transactions on Geoscience and Remote Sensing}, 2023.

\bibitem{zhao2023fast}
X.~Zhao, W.~Ding, Y.~An, Y.~Du, T.~Yu, M.~Li, M.~Tang, and J.~Wang, ``Fast
  segment anything,'' \emph{arXiv preprint arXiv:2306.12156}, 2023.

\bibitem{bokhovkin2019boundary}
A.~Bokhovkin and E.~Burnaev, ``Boundary loss for remote sensing imagery
  semantic segmentation,'' in \emph{International Symposium on Neural
  Networks}.\hskip 1em plus 0.5em minus 0.4em\relax Springer, 2019, pp.
  388--401.

\bibitem{markus2014use}
I.~Markus~Gerke, ``Use of the stair vision library within the {ISPRS} {2D}
  semantic labeling benchmark ({Vaihingen}),'' \emph{Use of the stair vision
  library within the isprs 2d semantic labeling benchmark (vaihingen)}, 2014.

\bibitem{zhang2023input}
Y.~Zhang, T.~Zhou, P.~Liang, and D.~Z. Chen, ``Input augmentation with {SAM}:
  Boosting medical image segmentation with segmentation foundation model,''
  \emph{arXiv preprint arXiv:2304.11332}, 2023.

\bibitem{li2023nnsam}
Y.~Li, B.~Jing, X.~Feng, Z.~Li, Y.~He, J.~Wang, and Y.~Zhang, ``{nnSAM}:
  Plug-and-play segment anything model improves {nnUNet} performance,''
  \emph{arXiv preprint arXiv:2309.16967}, 2023.

\bibitem{jiang2023segment}
P.-T. Jiang and Y.~Yang, ``Segment anything is a good pseudo-label generator
  for weakly supervised semantic segmentation,'' \emph{arXiv preprint
  arXiv:2305.01275}, 2023.

\bibitem{huang2023push}
Z.~Huang, H.~Liu, H.~Zhang, F.~Xing, A.~Laine, E.~Angelini, C.~Hendon, and
  Y.~Gan, ``Push the boundary of {SAM}: A pseudo-label correction framework for
  medical segmentation,'' \emph{arXiv preprint arXiv:2308.00883}, 2023.

\bibitem{li2019deep}
J.~Li, X.~Huang, and J.~Gong, ``Deep neural network for remote-sensing image
  interpretation: Status and perspectives,'' \emph{National Science Review},
  vol.~6, no.~6, pp. 1082--1086, 2019.

\bibitem{zheng2020foreground}
Z.~Zheng, Y.~Zhong, J.~Wang, and A.~Ma, ``Foreground-aware relation network for
  geospatial object segmentation in high spatial resolution remote sensing
  imagery,'' in \emph{Proceedings of the IEEE/CVF conference on computer vision
  and pattern recognition}, 2020, pp. 4096--4105.

\bibitem{duro2012comparison}
D.~C. Duro, S.~E. Franklin, and M.~G. Dub{\'e}, ``A comparison of pixel-based
  and object-based image analysis with selected machine learning algorithms for
  the classification of agricultural landscapes using spot-5 hrg imagery,''
  \emph{Remote sensing of environment}, vol. 118, pp. 259--272, 2012.

\bibitem{zhang2018object}
C.~Zhang, I.~Sargent, X.~Pan, H.~Li, A.~Gardiner, J.~Hare, and P.~M. Atkinson,
  ``An object-based convolutional neural network ({OCNN}) for urban land use
  classification,'' \emph{Remote sensing of environment}, vol. 216, pp. 57--70,
  2018.

\bibitem{martins2020exploring}
V.~S. Martins, A.~L. Kaleita, B.~K. Gelder, H.~L. da~Silveira, and C.~A. Abe,
  ``Exploring multiscale object-based convolutional neural network (multi-ocnn)
  for remote sensing image classification at high spatial resolution,''
  \emph{ISPRS Journal of Photogrammetry and Remote Sensing}, vol. 168, pp.
  56--73, 2020.

\bibitem{zhang2021style}
X.~Zhang, W.~Yu, M.-O. Pun, and M.~Liu, ``Style transformation-based change
  detection using adversarial learning with object boundary constraints,'' in
  \emph{2021 IEEE International Geoscience and Remote Sensing Symposium
  IGARSS}.\hskip 1em plus 0.5em minus 0.4em\relax IEEE, 2021, pp. 3117--3120.

\bibitem{li2022temporal}
H.~Li, Y.~Tian, C.~Zhang, S.~Zhang, and P.~M. Atkinson, ``Temporal sequence
  object-based cnn ({TS-OCNN}) for crop classification from fine resolution
  remote sensing image time-series,'' \emph{The Crop Journal}, vol.~10, no.~5,
  pp. 1507--1516, 2022.

\bibitem{rittenhouse2022object}
C.~D. Rittenhouse, E.~H. Berlin, N.~Mikle, S.~Qiu, D.~Riordan, and Z.~Zhu, ``An
  object-based approach to map young forest and shrubland vegetation based on
  multi-source remote sensing data,'' \emph{Remote Sensing}, vol.~14, no.~5, p.
  1091, 2022.

\bibitem{mi2020superpixel}
L.~Mi and Z.~Chen, ``Superpixel-enhanced deep neural forest for remote sensing
  image semantic segmentation,'' \emph{ISPRS Journal of Photogrammetry and
  Remote Sensing}, vol. 159, pp. 140--152, 2020.

\bibitem{zhang2021escnet}
H.~Zhang, M.~Lin, G.~Yang, and L.~Zhang, ``Escnet: An end-to-end
  superpixel-enhanced change detection network for very-high-resolution remote
  sensing images,'' \emph{IEEE Transactions on Neural Networks and Learning
  Systems}, 2021.

\bibitem{zhang2021object}
X.~Zhang, X.~Tan, G.~Chen, K.~Zhu, P.~Liao, and T.~Wang, ``Object-based
  classification framework of remote sensing images with graph convolutional
  networks,'' \emph{IEEE Geoscience and Remote Sensing Letters}, vol.~19, pp.
  1--5, 2021.

\bibitem{henaff2015deep}
M.~Henaff, J.~Bruna, and Y.~LeCun, ``Deep convolutional networks on
  graph-structured data,'' \emph{arXiv preprint arXiv:1506.05163}, 2015.

\bibitem{pan2021simplified}
X.~Pan, C.~Zhang, J.~Xu, and J.~Zhao, ``Simplified object-based deep neural
  network for very high resolution remote sensing image classification,''
  \emph{ISPRS Journal of Photogrammetry and Remote Sensing}, vol. 181, pp.
  218--237, 2021.

\bibitem{marmanis2018classification}
D.~Marmanis, K.~Schindler, J.~D. Wegner, S.~Galliani, M.~Datcu, and U.~Stilla,
  ``Classification with an edge: Improving semantic image segmentation with
  boundary detection,'' \emph{ISPRS Journal of Photogrammetry and Remote
  Sensing}, vol. 135, pp. 158--172, 2018.

\bibitem{liu2018ern}
S.~Liu, W.~Ding, C.~Liu, Y.~Liu, Y.~Wang, and H.~Li, ``{ERN}: Edge loss
  reinforced semantic segmentation network for remote sensing images,''
  \emph{Remote Sensing}, vol.~10, no.~9, p. 1339, 2018.

\bibitem{li2021abcnet}
R.~Li, S.~Zheng, C.~Zhang, C.~Duan, L.~Wang, and P.~M. Atkinson, ``Abcnet:
  Attentive bilateral contextual network for efficient semantic segmentation of
  fine-resolution remotely sensed imagery,'' \emph{ISPRS journal of
  photogrammetry and remote sensing}, vol. 181, pp. 84--98, 2021.

\bibitem{wang2021loveda}
J.~Wang, Z.~Zheng, A.~Ma, X.~Lu, and Y.~Zhong, ``{LoveDA}: A remote sensing
  land-cover dataset for domain adaptive semantic segmentation,'' in
  \emph{Proceedings of the Neural Information Processing Systems Track on
  Datasets and Benchmarks}, vol.~1, 2021, pp. 1--17.

\bibitem{audebert2018beyond}
N.~Audebert, B.~Le~Saux, and S.~Lef{\`e}vre, ``Beyond {RGB}: Very high
  resolution urban remote sensing with multimodal deep networks,'' \emph{ISPRS
  journal of photogrammetry and remote sensing}, vol. 140, pp. 20--32, 2018.

\bibitem{ma2022crossmodal}
X.~Ma, X.~Zhang, and M.-O. Pun, ``A crossmodal multiscale fusion network for
  semantic segmentation of remote sensing data,'' \emph{IEEE Journal of
  Selected Topics in Applied Earth Observations and Remote Sensing}, vol.~15,
  pp. 3463--3474, 2022.

\bibitem{liu2021swin}
Z.~Liu, Y.~Lin, Y.~Cao, H.~Hu, Y.~Wei, Z.~Zhang, S.~Lin, and B.~Guo, ``Swin
  transformer: Hierarchical vision transformer using shifted windows,'' in
  \emph{Proceedings of the IEEE/CVF international conference on computer
  vision}, 2021, pp. 10\,012--10\,022.

\end{thebibliography}
\end{document}